\documentclass{article}

\usepackage{nac_preprint}

\usepackage[utf8]{inputenc} 
\usepackage[T1]{fontenc}    
\usepackage{hyperref}       
\usepackage{url}            
\usepackage{booktabs}       
\usepackage{amsfonts}       
\usepackage{nicefrac}       
\usepackage{microtype}      
\usepackage{mathtools}
\usepackage{amsthm}
\usepackage{amsfonts}
\usepackage{wrapfig}
\usepackage{algorithm}
\usepackage{algorithmicx}
\usepackage[noend]{algpseudocode}
\algnewcommand{\LineComment}[1]{\State \(//\) #1}
\usepackage{comment}
\usepackage{soul}
\usepackage{enumitem}
\usepackage[dvipsnames]{xcolor}

\title{Improving Label Quality by Jointly Modeling\\ Items and Annotators}

\author{%
  Tharindu Cyril Weerasoriya
  \\
  Department of Computer Science\\
  Rochester Institute of Technology\\
  Rochester, NY, 14623 \\
  \texttt{cyriltcw@gmail.com} \\
   \And
   Alexander G. Ororbia \\
   Department of Computer Science\\
   Rochester Institute of Technology\\
   Rochester, NY, 14623 \\
   \texttt{ago@cs.rit.edu} \\
   \AND
   Christopher M. Homan \\
   Department of Computer Science\\
   Rochester Institute of Technology\\
   \texttt{cmh@cs.rit.edu} \\
}

\begin{document}

\maketitle

\begin{abstract}
  We propose a fully Bayesian framework
  for learning ground truth labels from noisy annotators.
  Our framework ensures scalability
  by factoring a generative, Bayesian soft clustering model over label distributions into the classic David and Skene joint annotator-data model. Earlier research along these lines has neither fully incorporated label distributions nor  explored clustering by annotators only or data only. Our framework incorporates all of these properties as:
  (1) a graphical model designed to provide better ground truth estimates of annotator responses as input to \emph{any} black box supervised learning algorithm, and 
  (2) a standalone neural model whose internal structure captures many of the properties of the graphical model.
  We conduct supervised learning experiments using both models and compare them to the performance of one baseline and a state-of-the-art model.
\end{abstract}

\section{Introduction}
\label{sec:intro}

The recent interest in few- and zero-shot learning and the re-emergence of weakly supervised learning speaks to the reality that ground truth labels are a limited resource and that, in many common situations, obtaining them remains a major challenge. Multiple sources estimate the the global costs of human annotators (only one of many sources of labels) to be approaching \$1--3 billion by $2026$ and growing \cite{caetz2019labels,kbv2020global}. Among the cost-driving challenges is the noise associated with many of the most common processes for obtaining labels. 

In this paper, we explore novel graphical and neural models that tie together two rather successful approaches, item-annotators tableaus \cite{dawid1979}, and label distribution learning (LDL) \cite{Geng2016}, based on converging studies in later research \cite{venanzi2014community,Liu2019HCOMP} on the use of clustering to boost the signal of noisy data. We adopt a theoretical framework motivated by the anthropologist Malinowski \cite{malinowski1967problem} and first used by Aroyo and Welty \cite{aroyo2014} in the context of machine learning to characterize meaning as a function of three components: 1) an act (represented by the learning task), 2) the symbols (the labels), and, 3) the referent (the annotators). Human labeling is a special challenge not only due to its great expense but also due to the fact that humans often disagree over the labels that they provide. In fact, it is precisely the problems where disagreement is most common that human input is hardest to replace through automation or sensing.



This paper specifically addresses the following research questions:
\begin{description}[leftmargin=0cm]
    \item[RQ1:] Do predictive graphical models for LDL that cluster on both item AND annotator distributions outperform those that do not?
    \item[RQ2:] Do predictive neural models for LDL outperform graphical models?
    \item[RQ3:] Do predictive neural models for LDL that cluster on both item AND annotator distributions outperform those that do not?
\end{description}
To help us answer these questions, we contribute two new models. The first is a generative graphical model that boosts conventional label distribution learning by clustering label distributions jointly in item and annotator label distribution spaces. Previous approaches have studied clustering in one space or the other. This is, to our knowledge, the first time that clustering has been applied simultaneously to both. Our second model is a neural-based adaptation of the graphical model. While the graphical model has a sound theoretical foundation, it is somewhat unwieldy from a computational perspective. The neural model sacrifices some rigor for more flexibility and algorithmic efficiency. 

\section{Related Work}
\label{sec:related_work}

The study of annotator disagreement has a long history, coincident with the emergence of data-driven behavioral research \cite{cohen1960coefficient}. 
As mentioned in the introduction, our approach factors a generative, fully Bayesian soft clustering model over label distributions into the classic David and Skene joint annotator-data model. We thus join two streams of loosely related research.

The earliest of these, by the aforementioned David and Skene \cite{dawid1979}, all use the multiple labels associated with each data item and each annotator to jointly estimate the (binary) ground truth label of each item and the error rate of each annotator, usually via an EM algorithm that alternates between estimates of ground truth (as hidden variables) and maximizing the error rate. 

Later researchers put this model on a fully Bayesian foundation \cite{raykar2010learning,kim2012bayesian}, and consider more complex models of annotator or ground truth, or both \cite{whitehill2009whose,northcutt2019confident}. Notably, (as spam is a common problem in crowdsourced label sets) several investigators distinguish between honest and dishonest annotators \cite{raykar2012eliminating,hovy-etal-2013-learning}. More recently, investigators have studied clustering as an unsupervised approach to discover annotators with similar behavior \cite{venanzi2014community}. Yet all of the these approaches are based on the assumption that each item is associated with a single ground truth label. As a consequence, the more advanced models tend to represent annotators as a distribution over a confusion matrix and data items as a single distribution (sometimes conditioned on input features).

Label distribution learning, by contrast, assumes that ground truth itself is a distribution.  However, the distribution does not necessarily come from a population of annotators \cite{Geng2016,Gao2017,wang2019theoretical,Zhang2020}. It has found application in a diverse range of settings \cite{Geng2014,Geng2015,Ren2017,Ling2018,shirani2019,Yang2020}.
Here, the goal to predict the distribution of labels associated with an item, rather than a single ground truth label. It is relatively natural, in this setting, to consider clustering together related data items to improve ground truth estimates of label distributions, as several researchers have done, either in the feature space of the items \cite{Zheng2018,Zeng2020} or directly in the label space of the items themselves \cite{Liu2019,Liu2019HCOMP,Weerasooriya2020}.


\section{Problem Statement}
\label{sec:ldl_problem_setup}
Let $\boldsymbol{X}$ be an $M$-element collection of (unlabeled) \emph{data items} and $\boldsymbol{Y} \in \mathbb{N}^{M \times N}$ a matrix of \emph{annotator labels} for some $N$, where each row of $\boldsymbol{Y}$ corresponds to a data item and each column to an \emph{annotator}. Ideally, we would regard each entry $\boldsymbol{Y}_{m,n}$ as a probably distribution over a set of labels $\{1,\ldots,P\}$ for some fixed $P$, where the distribution represents uncertainty about what label annotator $n$ would provide to item $m$. But here we simplify the model under the assumption that each annotator either provides a single label or none at all.

So for our purposes, $\boldsymbol{Y}$ is a sparse matrix, where $\boldsymbol{Y}_{m,n} \in \{0,\ldots,P\}$ and $\boldsymbol{Y}_{m,n} = 0$ indicates that annotator $n$ did not label item $m$. Crucially, we assume that each annotator \emph{could} label the item if asked; however, we have no information about that particular annotator.  As this is a sparse matrix, it is convenient to let $A = \{(m,n)~|~\boldsymbol{Y}_{m,n} \neq 0\}$ and $A_p = \{(m,n)~|~\boldsymbol{Y}_{m,n} = p\}$.

We consider two gold standards: $f_{\rm dist}$ and $f_{\rm max}$, defined for data item $\boldsymbol{X}_m$ as $f_{\rm dist}(\boldsymbol{X}_m) =_{\rm def} \mathrm{P}(p=Y_{m,n}~|~m, Y_{m,n} > 0)$, for $m,n$ chosen uniformly at random and $f_{\rm max}(\boldsymbol{X}_m) =_{\rm def}  \arg\max_p \mathrm{P}(p=Y_{m,n}~|~m, Y_{m,n} > 0)$. In other words, $f_{\rm dist}(\boldsymbol{X}_m)$ represents the gold standard \emph{label distribution} associated with each data item and $f_{\rm max}$ is the gold standard single label that is most likely, according to $f_{\rm dist}$. Note that $f_{\rm max}$ is more commonly used than $f_{\rm dist}$.

Our learning goals, then, are to produce hypotheses $h_{\rm dist}$ and $h_{\rm max}$ that approximate $f_{\rm dist}$ and $f_{\rm max}$, respectively, given $\boldsymbol{X}$ and $\boldsymbol{Y}$. Most learning settings tacitly assume that annotator disagreement is a sign of \emph{noise} or \emph{error} and ignore $d_{\rm dist}$ entirely. \emph{Label distribution learning} does the opposite: it assumes that annotator disagreement is \emph{meaningful} and specifically seeks to minimize the loss between $h_{\rm dist}$ and $f_{\rm dist}$. Obviously, both approaches rely on extreme assumptions that, in practice, are never entirely true. However, research has shown that even when $f_{\rm max}$ is the goal learning $h_{\rm dist}$ then taking $h_{\rm max}(\boldsymbol{X}_m) =_{\rm def} \arg\max_p \mathrm{P}(h_{\rm dist}(\boldsymbol{X}_m) = p)$ often provides better results than learning $h_{\rm max}$ directly \cite{venanzi2014community,Liu2019HCOMP,Weerasooriya2020}, and this is what we do here.

\section{The Probabilistic Graph Model}
\label{sec:pgm}
We call $f_{\rm dist}$ and $f_{\rm max}$ gold standards, not ground truth, because of the sparseness of $\boldsymbol{Y}$.  Although several researchers have shown that, for the purpose estimating  $f_{\rm max}$, three to ten annotators is sufficient \cite{callison2009fast, denkowski2010exploring}, those numbers are far too small to provide reliable samples of the true distributions of annotator opinions.  In this section, we introduce a new graphical model  estimating the ground truth label distribution, i.e., the distribution of labels from the entire population of annotators, of each item (which we normally do not have). This model is based on the assumptions that: (1) all data items (respectively, annotators) are drawn from one of $K$ (respectively, $L$) latent classes\footnote{Hereafter, to reduce confusion, we reserve ``class'' to refer only to the different label choices, as they typically represent an observable class to which the data item belongs, even though the idea of labels as indivisible classes runs contrary to the spirit of LDL.} or \emph{clusters}, (2) the label distribution for each item is strictly a function of the cluster to which it belongs, (3) the sample of labels given for each item is strictly a function of the distribution of the cluster to which each annotator belongs, and (4) the items and annotators are identically and independently sampled (i.i.d.) and  matched uniformly at random. 

\begin{figure}
    \centering
    \includegraphics[width=\textwidth]{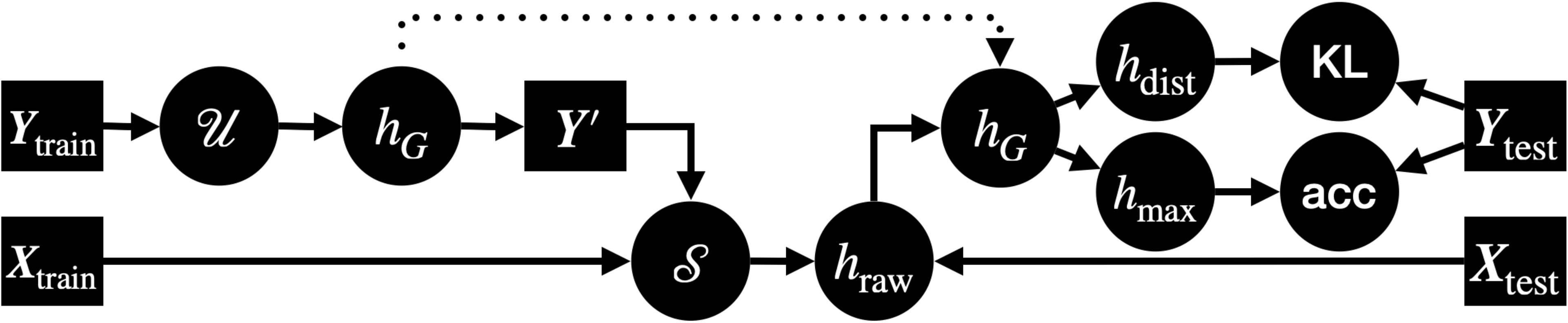}
    \caption{This workflow diagram shows the dual roles of the graphical model $h_{G}$, as the output of a supervised learning process $\mathcal{U}$ on the training labels, which is used to improve the ground truth estimations $\boldsymbol{Y}'$ of the gold standard training label distributions $\boldsymbol{Y}_{\rm train}$ for supervised learning $\mathcal{S}$, and, once the model $h_{\rm raw}$ is learned, as a post-processing step after prediction to generate final hypotheses $h_{\rm dist}$ and $h_{\rm max}$. We use as our evaluation metrics the accuracy on the most likely label for single label prediction $h_{\rm max}$ and KL divergence for label distribution learning $h_{\rm dist}$.}
    \label{fig:workflow_graph}
\end{figure}

We then use the graphical model $h_G$ to guide supervised learning as a means of data regularization. See Figure \ref{fig:workflow_graph}. We first use it as a preprocessing step to supervised learning on our label matrix $Y$, by snapping each input label distribution $\boldsymbol{Y}_m$ to the generating distribution of the most likely item cluster before training. Note that any supervised learning method can work as long as it can use a distribution of labels and the supervising signal. For instance, in our experiments (see Section \ref{sec:experiments}) we use a combination of deep language models and simple dense networks.
Next, after the predictive model $h_{\rm dist}$ is learned, we postprocess each prediction by, snapping each output $h_{\rm dist}(\boldsymbol{X}_m)$ again to the most likely item cluster.

\begin{figure}
\includegraphics[width=\textwidth]{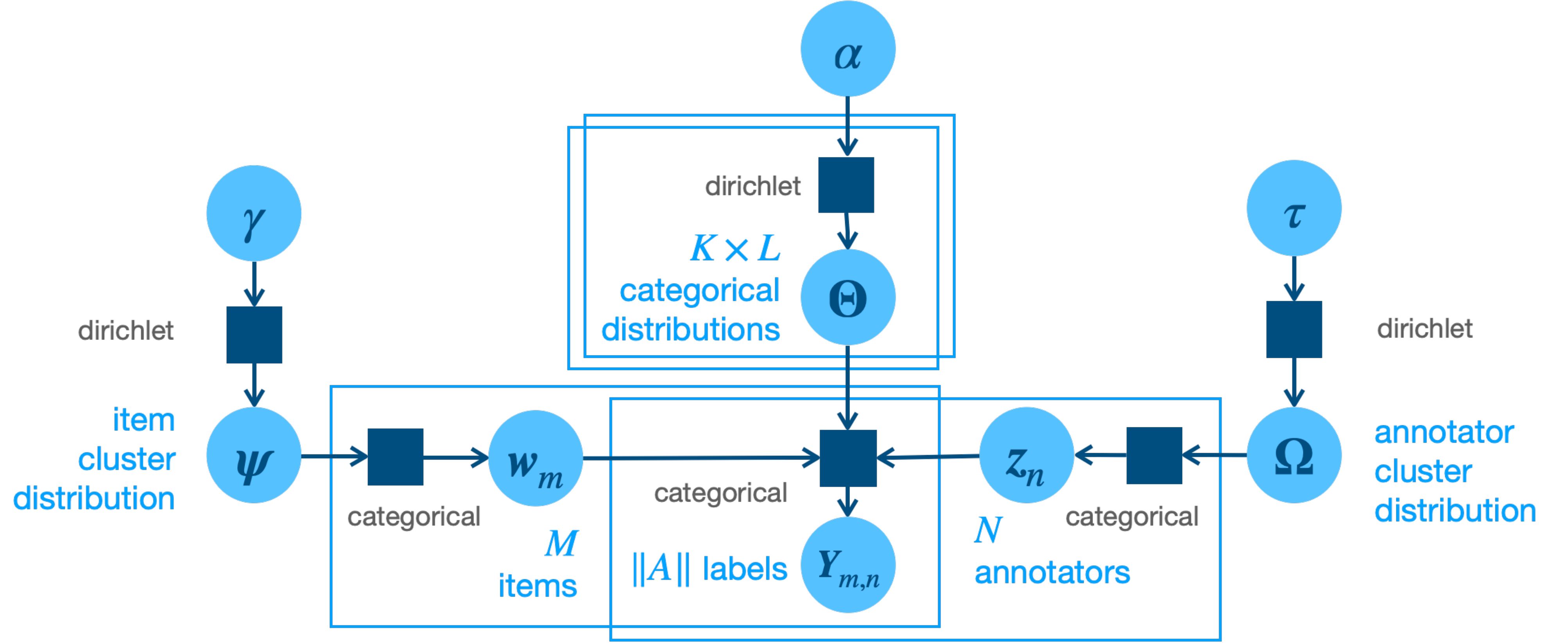}
\caption{Plate diagram for the generative graph model.}
\label{fig:plate_diagram}
\end{figure}

 Algorithm \ref{algo:graph_process} describes  the model from a generative perspective. See also Figure \ref{fig:plate_diagram}. In addition to the numbers of item and annotator clusters $K$ and $L$, the model takes three hyperparameters, $\alpha \in \mathbb{R}^P$ (recall that $P$ is the number of label classes), $\gamma \in \mathbb{R}^K$, and $\tau \in \mathbb{R}^L$, each of which represents a Dirichlet prior on a categorical distribution. It produces $\boldsymbol{\Theta}_{k,l}$, the label distribution for each item cluster $k$ and annotator cluster $l$, $\boldsymbol{\psi}$, the marginal class distribution of items, $\boldsymbol{\Omega}$, the marginal class distribution of annotators, $\boldsymbol{w}_n$ is the hidden variable representing the class of item $m$, and $\boldsymbol{z}_n$, the hidden variable representing the class of annotator $n$. Each of these objects is a categorical distribution, and so for convenience we use subscripts to indicate individual categorical probabilities. E.g., $\boldsymbol{\Theta}_{k,l,p} = \mathrm{P}(\mbox{The category is }p)$ and $\boldsymbol{\Omega}_l = \mathrm{P}(\mbox{The category is }l)$.

\begin{algorithm}[!t]
\caption{The generative process for $h_G$.}
\label{algo:graph_process}
\begin{algorithmic}
   \State {\bfseries Input:} Integers  $K$, $L$, $M$, $N$, and $P$; Dirichlet hyperparameters $\alpha \in \mathbb{R}^P$, $\gamma \in \mathbb{R}^K$, and $\tau \in \mathbb{R}^L$, assignments $A \subseteq \{1,\ldots M\} \times \{1,\ldots N\}$
   \Function{GenGraph}{$K$, $L$, $M$, $N$, $P$, $\alpha$, $\gamma$, and $\tau$} 
        \State Choose $\boldsymbol{\Theta} \sim {\rm Dir}_P(\alpha)^{K \times L}$, \Comment{One distribution for each item/annotator cluster pair $(k,l)$}
        \State Choose $\boldsymbol{\psi} \sim {\rm Dir}_K(\gamma)$, \Comment{Distribution of item clusters}
        \State Choose $\boldsymbol{\Omega} \sim {\rm Dir}_L(\tau)$, \Comment{Distribution of annotator clusters}
        \State Choose $\boldsymbol{w}\sim {\rm Cat}_K(\boldsymbol{\psi})^M$, \Comment{Assign one latent cluster to each item}
        \State Choose $\boldsymbol{z} \sim {\rm Cat}_L(\boldsymbol{\Omega})^N$, \Comment{Assign one latent cluster to each annotator}
        \State Choose $\boldsymbol{Y} \sim \bigtimes_{(m,n) \in \boldsymbol{A}} {\rm Cat}_P(\boldsymbol{\Theta}_{\boldsymbol{w}_m, \boldsymbol{z}_n})$. \Comment{Assign labels according to each annotator, item assignment}
    \EndFunction
\end{algorithmic}
\end{algorithm}

The joint distribution, under the independence assumptions implied in the algorithm above, is
\begin{align*}
 \MoveEqLeft
\mathrm{P}_{\alpha,\gamma,\tau}\left(\boldsymbol{\Theta},\boldsymbol{\psi}, \boldsymbol{w}, \boldsymbol{\Omega}, \boldsymbol{z}, \boldsymbol{Y} ~|~ \boldsymbol{A}\right)\\
&=\mathrm{P}_{\alpha}(\boldsymbol{\Theta})\mathrm{P}_{\gamma}(\boldsymbol{\psi}) \mathrm{P}(\boldsymbol{w}|\boldsymbol{\psi})\mathrm{P}_{\tau}(\boldsymbol{\Omega})\mathrm{P}(\boldsymbol{z}|\boldsymbol{\Omega})\prod_{(m,n) \in \boldsymbol{A}}\mathrm{P}(\boldsymbol{Y}_{m,n}|\boldsymbol{\Theta}_{\boldsymbol{w}_m,\boldsymbol{z}_n})\\
&=\mathrm{P}_{\alpha}(\boldsymbol{\Theta})\mathrm{P}_{\gamma}(\boldsymbol{\psi}) \prod_{m} \mathrm{P}(\boldsymbol{w}_m|\boldsymbol{\psi})\mathrm{P}_{\tau}(\boldsymbol{\Omega})\prod_{n}\mathrm{P}(\boldsymbol{z}_n|\boldsymbol{\Omega})\prod_{(m,n) \in \boldsymbol{A}}\mathrm{P}(\boldsymbol{Y}_{m,n}|\boldsymbol{\Theta}_{\boldsymbol{w}_m,\boldsymbol{z}_n})
\end{align*}
Note that our distributions are conditioned on $\boldsymbol{A}$. That is, we always know beforehand which annotators are assigned to which items. Unfortunately, the coupling between items and annotators makes exact inference hard, and even resistant to variational approximation. It is, however, relatively easy to perform simulated annealing over the parameters 
 $\boldsymbol{\Theta}$, $\boldsymbol{\psi}$, $\boldsymbol{\Omega}$ and hidden variables $\boldsymbol{w}$, and $\boldsymbol{z}$ as well as EM, using belief propagation to estimate probability distributions of $\boldsymbol{w}$ and $\boldsymbol{z}$ 
during the expectation phase. We explore both learning algorithms here.

We now describe in more detail how we use the model. We partition our data into training $(\boldsymbol{X}_{\rm train}, \boldsymbol{Y}_{\rm train})$, development $(\boldsymbol{X}_{\rm dev}, \boldsymbol{Y}_{\rm dev})$, and test $(\boldsymbol{X}_{\rm test}, \boldsymbol{Y}_{\rm test})$ splits. During training, we first apply one of our two unsupervised learning algorithms $h_G = \mathcal{U}(\boldsymbol{Y}_{\rm train})$ to learn a graphical model $h_G = (\boldsymbol{\Theta}, \boldsymbol{\psi}, \boldsymbol{\Omega},\boldsymbol{w}, \boldsymbol{z})$ from $\boldsymbol{Y}_{\rm train}$. Note that this provides estimates $\boldsymbol{w}$ of the latent item cluster to which each item belongs (simulated annealing provides a hard clustering, EM a soft clustering, but with EM we consider only the most likely cluster). Then, before supervised learning, we replace row $\boldsymbol{Y}_{{\rm test},m}$ with the marginal label distribution associated with item cluster $\boldsymbol{w}_m$,
\begin{align}
\label{eq:infer}
    \boldsymbol{Y}'_{m} = \sum_l \boldsymbol{\Omega}_l \boldsymbol{\Theta}_{\boldsymbol{w}_ml}, 
\end{align}
And perform supervised learning $h_{\rm raw} = \mathcal{S}(\boldsymbol{X}_{\rm train},\boldsymbol{Y}')$, yielding a raw label distribution learning predictor. Note that $\boldsymbol{Y}'$ is not a matrix of annotator labels, as $\boldsymbol{Y}_{\rm train}$ is, but a vector of probability distributions over labels.

For inference AFTER training (i.e., we do not perform this step during training), for any input $\boldsymbol{x}$ we project the output of $h_{\rm raw}(\boldsymbol{x})$ onto our graph model $h_G$ to predict the item cluster membership of item $x$, i.e., let $\mathrm{w}(\boldsymbol{x})$ denote a random variable for the item cluster assignment of $\boldsymbol{x}$. Then
\begin{align}
    \mathrm{P}(\mathrm{w}(\boldsymbol{x}) = k) \sim \sum_l \boldsymbol{\psi}_k \boldsymbol{\Omega}_l  \mathrm{P}\left(h_{\rm raw}(\boldsymbol{x}) \sim {\rm Cat}_P(\boldsymbol{\Theta}_{k,l})\right)
\end{align}
We then assign to $x$ the item cluster $\arg\max_k \mathrm{P}(\mathrm{w}(\boldsymbol{x}) = k)$, using Equation (\ref{eq:infer}) to compute $h_{\rm dist}(\boldsymbol{x})$ and define $h_{\rm max}(\boldsymbol{x}) =_{\rm def} \arg\max_p \mathrm{P} \left(h_{\rm dist}(\boldsymbol{x}) = p\right)$.

\subsection{Equations for the Graphical Model}
Both the simulated annealing and EM algorithms studied here are based on the log likelihood formulation of the graphical model:

\begin{align}
    \label{eq:graphloglikely}
     \MoveEqLeft
\ell_{\alpha,\gamma,\tau}\left(\boldsymbol{\Theta},\boldsymbol{\psi}, \boldsymbol{w}, \boldsymbol{\Omega}, \boldsymbol{z}, \boldsymbol{Y} ~|~ \boldsymbol{A}\right)\notag\\
&=\sum_{k,l} \log p_{\alpha}(\boldsymbol{\Theta}_{k,l})  + \log p_{\gamma}(\boldsymbol{\psi})+ \sum_{m} \log p(\boldsymbol{w}_m | \boldsymbol{\psi})\notag \\
&\hspace{.25cm} +p_{\tau}(\boldsymbol{\Omega}) + \sum_{n}\log p(\boldsymbol{z}_n|\boldsymbol{\Omega}) + \sum_{(m,n) \in \boldsymbol{A}} \log p(\boldsymbol{Y}_{m,n}|\boldsymbol{\Theta}_{\boldsymbol{w}_m,\boldsymbol{z}_n})\notag\\
&=(\alpha - 1)\sum_{k,l,p} \log \boldsymbol{\Theta}_{k,l,p} 
+  (\gamma - 1)\sum_k \log \boldsymbol{\psi}_k + \sum_{m} \log \boldsymbol{\psi}_{\boldsymbol{w}_m} \notag\\
&\hspace{.25cm}+ (\tau - 1) \sum_{l} \log \boldsymbol{\Omega}_l  + \sum_{n}\log \boldsymbol{\Omega}_{\boldsymbol{z}_n}
+ \sum_{(m,n) \in \boldsymbol{A}} \log \boldsymbol{\Theta}_{\boldsymbol{w}_m,\boldsymbol{z}_n,\boldsymbol{Y}_{m,n}}.
\end{align}

\subsubsection{Simulated Annealing Equations}
Algorithm \ref{alg:sa} shows pseudocdoe for performing simulated annealing on the graph model, where the update probabilities are derived from the log-likelihood function as follows (note that we iterate over both the parameters and hidden variables).
\begin{algorithm}[!t]
\caption{The simulated annealing process for $h_G$.}
\label{alg:sa}
\begin{algorithmic}
   \State {\bfseries Input:} Integers  $K$, $L$, $M$, $N$, and $P$; Dirichlet hyperparameters $\alpha \in \mathbb{R}^P$, $\gamma \in \mathbb{R}^K$, and $\tau \in \mathbb{R}^L$, assignments $A \subseteq \{1,\ldots M\} \times \{1,\ldots N\}$
   \Function{SAGraph}{$K$, $L$, $M$, $N$, $P$, $\alpha$, $\gamma$, and $\tau$, $A$} 
   \State Randomly initialize $\boldsymbol{\Theta}$, $\boldsymbol{\psi}$, $\boldsymbol{w}$, $\boldsymbol{\Omega}$, $\boldsymbol{z}$
   \For{$t \gets 0, 1, \ldots$} \Comment{Repeat until convergence}
   \State Compute $\textbf{P}_{\alpha, \gamma, \tau}(\boldsymbol{\Theta}, \boldsymbol{\psi}, \boldsymbol{w}, \boldsymbol{\Omega}, \boldsymbol{z}~|~A)$
   \State Store parameters in a priority queue keyed on $\textbf{P}_{\alpha, \gamma, \tau}(\boldsymbol{\Theta}, \boldsymbol{\psi}, \boldsymbol{w}, \boldsymbol{\Omega}, \boldsymbol{z}~|~A)$
   \For{$k \gets 1, \ldots, K$} 
   \For{$l \gets 1, \ldots, L$}
   \State Sample a new $\boldsymbol{\Theta}^{\rm new}_{k,l}$
   \If{log-prob w/ $\boldsymbol{\Theta}^{\rm new}_{k,l} > \boldsymbol{\Theta}^{\rm old}_{k,l}$}
   \State $\boldsymbol{\Theta}_{k,l} \leftarrow \boldsymbol{\Theta}^{\rm new}_{k,l}$
   \Else \State $\boldsymbol{\Theta}_{k,l} \leftarrow \boldsymbol{\Theta}^{\rm new}_{k,l}$ with prob $e^{\delta_{\boldsymbol{\Theta}_{i,j}}(t+1)}$
   \EndIf
   \EndFor
   \EndFor
   \State Sample a new $\boldsymbol{\psi}^{\rm new}$
   \If{log-prob w/ $\boldsymbol{\psi}^{\rm new} > \boldsymbol{\psi}^{\rm old}$}
   \State $\boldsymbol{\psi} \gets \boldsymbol{\psi}^{\rm new}$
   \Else \State $\boldsymbol{\psi} \gets \boldsymbol{\psi}^{\rm new}$ with prob $e^{\delta_{\boldsymbol{\psi}}(t+1)}$
   \EndIf
   \State Sample a new $\boldsymbol{\Omega}^{\rm new}$
   \If{log-prob w/ $\boldsymbol{\Omega}^{\rm new} > \boldsymbol{\Omega}^{\rm old}$}
   \State $\boldsymbol{\Omega} \leftarrow \boldsymbol{\Omega}^{\rm new}$
   \Else \State $\boldsymbol{\Omega} \leftarrow \boldsymbol{\Omega}^{\rm new}$ with prob $e^{\delta_{\boldsymbol{\Omega}}(t+1)}$
   \EndIf
   \For{$n \in 1, \ldots, M$}
   \State Sample a new $\boldsymbol{w}^{\rm new}_m$
   \If{log-prob w/ $\boldsymbol{w}^{\rm new}_m > \boldsymbol{w}^{\rm old}_m$}
   \State $\boldsymbol{w}_m\leftarrow \boldsymbol{w}^{\rm new}_m$
   \Else $\boldsymbol{w}_m\leftarrow \boldsymbol{w}^{\rm new}_m$ with prob $e^{\delta_{\boldsymbol{w}_m}(t+1)}$
   \EndIf
   \EndFor
   \For{$n \in N$}
   \State Sample a new $\boldsymbol{z}^{\rm new}_n$
   \If{the log-prob w/ $\boldsymbol{z}^{\rm new}_n > \boldsymbol{z}^{\rm old}_n$}
   \State $\boldsymbol{z}_n\leftarrow \boldsymbol{z}^{\rm new}_n$
   \Else \State $\boldsymbol{z}_n\leftarrow \boldsymbol{z}^{\rm new}_n$ with prob $e^{\delta_{\boldsymbol{z}_n}(t+1)}$
   \EndIf
   \EndFor
   \EndFor
   \Return the parameters that maximize $\textbf{P}_{\alpha, \gamma, \tau}(\boldsymbol{\Theta}, \boldsymbol{\psi}, \boldsymbol{w}, \boldsymbol{\Omega}, \boldsymbol{z}~|~A)$.
    \EndFunction
\end{algorithmic}
\end{algorithm}
\begin{align*}
 \MoveEqLeft
\delta_{\boldsymbol{\Theta}_{i,j}} = \ell_{\alpha,\gamma,\tau}\left(\boldsymbol{\Theta}^{\rm NEW
}_{k,l}~|~\boldsymbol{\Theta}_{-k,-l}, \boldsymbol{\psi}, \boldsymbol{w}, \boldsymbol{\Omega}, \boldsymbol{z}, \boldsymbol{Y},  \boldsymbol{A}\right) - \ell_{\alpha,\gamma,\tau}\left(\boldsymbol{\Theta}^{\rm OLD}_{k,l}~|~\boldsymbol{\Theta}_{-k,-l}, \boldsymbol{\psi}, \boldsymbol{w}, \boldsymbol{\Omega}, \boldsymbol{z}, \boldsymbol{Y},  \boldsymbol{A}\right)\\ 
&=(\alpha - 1)\sum_{p} \log \frac{\boldsymbol{\Theta}^{\rm NEW}_{k,l,p}}{\boldsymbol{\Theta}^{\rm OLD}_{k,l,p}} + \sum_{(m, n) \in \boldsymbol{A}: (\boldsymbol{w}_m,\boldsymbol{z}_n) = (k,l)} \log \frac{\boldsymbol{\Theta}^{\rm NEW}_{k,l,\boldsymbol{Y}_{m,n}
}}{\boldsymbol{\Theta}^{\rm OLD}_{k,l,\boldsymbol{Y}_{m,n}}}\\
\\
\MoveEqLeft
\delta_{\boldsymbol{\psi}} = \ell_{\alpha,\gamma,\tau}\left(\boldsymbol{\psi}^{\rm NEW
}~|~\boldsymbol{\Theta}, \boldsymbol{w}, \boldsymbol{\Omega}, \boldsymbol{z}, \boldsymbol{Y},  \boldsymbol{A}\right) - \ell_{\alpha,\gamma,\tau}\left(\boldsymbol{\psi}^{\rm OLD}~|~\boldsymbol{\Theta},  \boldsymbol{w}, \boldsymbol{\Omega}, \boldsymbol{z}, \boldsymbol{Y},  \boldsymbol{A}\right)\\ 
&=(\gamma - 1) \sum_{k} \log \frac{\boldsymbol{\psi}^{\rm NEW}_k}{\boldsymbol{\psi}^{\rm OLD}_k}\\
\MoveEqLeft
\delta_{\boldsymbol{\Omega}} = \ell_{\alpha,\gamma,\tau}\left(\boldsymbol{\Omega}^{\rm NEW
}~|~\boldsymbol{\Theta}, \boldsymbol{\psi}, \boldsymbol{w}, \boldsymbol{z}, \boldsymbol{Y},  \boldsymbol{A}\right) - \ell_{\alpha,\gamma,\tau}\left(\boldsymbol{\Omega}^{\rm OLD}~|~\boldsymbol{\Theta}, \boldsymbol{\psi}, \boldsymbol{w}, \ \boldsymbol{z}, \boldsymbol{Y},  \boldsymbol{A}\right)\\ 
&=(\tau - 1) \sum_{l} \log \frac{\boldsymbol{\Omega}^{\rm NEW}_l}{\boldsymbol{\Omega}^{\rm OLD}_l}\\
\end{align*}
\begin{align*}
\MoveEqLeft
\delta_{\boldsymbol{w}_m} = \ell_{\alpha,\gamma,\tau}\left(\boldsymbol{w}^{\rm NEW
}_{m}~|~\boldsymbol{\Theta}, \boldsymbol{\psi},\boldsymbol{w}_{-m},\boldsymbol{\Omega}, \boldsymbol{z}, \boldsymbol{Y},  \boldsymbol{A}\right) - \ell_{\alpha,\gamma,\tau}\left(\boldsymbol{w}^{\rm OLD
}_{m}~|~\boldsymbol{\Theta}, \boldsymbol{\psi},\boldsymbol{w}_{-m},\boldsymbol{\Omega}, \boldsymbol{z}, \boldsymbol{Y},  \boldsymbol{A}\right)\\
    &= \log \frac{\boldsymbol{\psi}_{\boldsymbol{w}^{\rm NEW}_m}}{\boldsymbol{\psi}_{\boldsymbol{w}^{\rm OLD}_m}} +  \sum_{n: (m,n) \in \boldsymbol{A}}  \log \frac{\boldsymbol{\Theta}_{\boldsymbol{w}^{\rm NEW}_m, \boldsymbol{z}_n,\boldsymbol{Y}_{m,n}} }{\boldsymbol{\Theta}_{\boldsymbol{w}^{\rm OLD}_m, \boldsymbol{z}_n,\boldsymbol{Y}_{m,n}}}\\
\MoveEqLeft
\delta_{\boldsymbol{z}_n} = \ell_{\alpha,\gamma,\tau}\left(\boldsymbol{z}^{\rm NEW
}_{m}~|~\boldsymbol{\Theta}, \boldsymbol{\psi},\boldsymbol{w},\boldsymbol{\Omega}, \boldsymbol{z}_{-m}, \boldsymbol{Y},  \boldsymbol{A}\right) - \ell_{\alpha,\gamma,\tau}\left(\boldsymbol{z}^{\rm OLD
}_{m}~|~\boldsymbol{\Theta}, \boldsymbol{\psi},\boldsymbol{w},\boldsymbol{\Omega}, \boldsymbol{z}_{-m}, \boldsymbol{Y},  \boldsymbol{A}\right)\\
&= \log \frac{\boldsymbol{\Omega}_{\boldsymbol{z}^{\rm NEW}_n}}{\boldsymbol{\Omega}_{\boldsymbol{z}^{\rm OLD}_n}} + \sum_{m: (m,n) \in \boldsymbol{A}} \log \frac{\boldsymbol{\Theta}_{\boldsymbol{w}_m, \boldsymbol{z}^{\rm NEW}_n,\boldsymbol{Y}_{m,n}} }{\boldsymbol{\Theta}_{\boldsymbol{w}_m, \boldsymbol{z}^{\rm OLD}_n,\boldsymbol{Y}_{m,n}}}
\end{align*}

\subsubsection{EM Equations}
The expectation formulation of equation (\ref{eq:graphloglikely}) is
\begin{align}
\MoveEqLeft
E_{\hat{\boldsymbol{w}},\hat{\boldsymbol{z}}}\left[\ell_{\alpha,\gamma,\tau}\left(\boldsymbol{\Theta},\boldsymbol{\psi}, \boldsymbol{w}, \boldsymbol{\Omega}, \boldsymbol{z}, \boldsymbol{Y} ~|~ \boldsymbol{A}\right)\right]\notag\\
&=\sum_{k,l} \log p_{\alpha}(\boldsymbol{\Theta}_{k,l})  + \log p_{\gamma}(\boldsymbol{\psi})+ \sum_{m}E_{\hat{\boldsymbol{w}},\hat{\boldsymbol{z}}}\left[ \log p(\boldsymbol{w}_m | \boldsymbol{\psi})\right]\notag \\
&\hspace{.25cm} +p_{\tau}(\boldsymbol{\Omega}) + \sum_{n}E_{\hat{\boldsymbol{w}},\hat{\boldsymbol{z}}}\left[\log p(\boldsymbol{z}_n|\boldsymbol{\Omega})\right] + \sum_{(m,n) \in \boldsymbol{A}} E_{\hat{\boldsymbol{w}},\hat{\boldsymbol{z}}}\left[\log  p(\boldsymbol{Y}_{m,n}|\boldsymbol{\Theta}_{\boldsymbol{w}_m,\boldsymbol{z}_n})\right]\notag\\
&=(\alpha - 1)\sum_{k,l,p} \log \boldsymbol{\Theta}_{k,l,p} 
+  (\gamma - 1)\sum_k \log \boldsymbol{\psi}_k + \sum_{m} \sum_k \boldsymbol{w}_{mk} \log \boldsymbol{\psi}_k \notag\\
&\hspace{.25cm}+ (\tau - 1) \sum_{l} \log \boldsymbol{\Omega}_l  + \sum_{n} \sum_l \boldsymbol{z}_{nl} \log \boldsymbol{\Omega}_l
+ \sum_{(m,n) \in \boldsymbol{A}} \sum_k \sum_l \boldsymbol{w}_{mk} \boldsymbol{z}_{nl} \log \boldsymbol{\Theta}_{k,l,\boldsymbol{Y}_{m,n}}.
\end{align}
Algorithm \ref{alg:em} summarizes the process. Note that we perform the ``M'' step before the ``E'' step in each round. This is arbitrary. Also, though we perform at most ten rounds of belief propagation at each outer iteration, in practice it we usually get convergence within two rounds, which we define as
\begin{align}
    \sum_{n} KL(\boldsymbol{w}^{\rm NEW}_n || \boldsymbol{w}^{\rm OLD}_n ) + \sum_{m} KL(\boldsymbol{z}^{\rm NEW}_m || \boldsymbol{z}^{\rm OLD}_m).
\end{align}

\begin{algorithm}[h]
\caption{The EM process for $h_G$.}
\label{alg:em}
\begin{algorithmic}
   \State {\bfseries Input:} Integers  $K$, $L$, $M$, $N$, and $P$; Dirichlet hyperparameters $\alpha \in \mathbb{R}^P$, $\gamma \in \mathbb{R}^K$, and $\tau \in \mathbb{R}^L$, assignments $A \subseteq \{1,\ldots M\} \times \{1,\ldots N\}$
   \Function{SAGraph}{$K$, $L$, $M$, $N$, $P$, $\alpha$, $\gamma$, and $\tau$, $A$} 
   \State Randomly initialize $\boldsymbol{\Theta}$, $\boldsymbol{\psi}$, $\boldsymbol{w}$, $\boldsymbol{\Omega}$, $\boldsymbol{z}$
   \Loop \Comment{Repeat until convergence}
   \State Compute $\boldsymbol{\Theta}$, $\boldsymbol{\psi}$, and $\boldsymbol{\Omega}$, according to the maximization equations given in Section \ref{sec:maxeq}.
      \Loop \Comment{Do ten times or until convergence}
     \State Update $\boldsymbol{w}$ according to the belief propagation equation given in Section \ref{sec:belief}
      \State Update $\boldsymbol{z}$ according to the belief propagation equation given in Section \ref{sec:belief}
    \EndLoop
   \EndLoop
   \Return $\boldsymbol{\Theta}, \boldsymbol{\psi}, \boldsymbol{w}, \boldsymbol{\Omega}, \boldsymbol{z}$.
    \EndFunction
\end{algorithmic}
\end{algorithm}
\subsubsection{Maximization Equations for Model Parameters}
\label{sec:maxeq}
Subject to the constraints 

\begin{align*}
   \boldsymbol{\Lambda}_{\boldsymbol{\Theta},\boldsymbol{\psi},\boldsymbol{\Omega}} &= \sum_{k,l}\lambda_{\boldsymbol{\Theta}_{k,l}}\left(1 - \sum_p \boldsymbol{\Theta}_{k,l,p}\right) +  \lambda_{\boldsymbol{\psi}}\left(1 - \sum_{k=1}^{K} \boldsymbol{\psi}_{k}\right) + \lambda_{\boldsymbol{\Omega}}\left(1 - \sum_{l=1}^{L} \boldsymbol{\Omega}_{l}\right),
\end{align*}
we can maximize the parameters $\boldsymbol{\Theta},\boldsymbol{\psi},\boldsymbol{\Omega}$ by take partial derivatives and setting to zero. Thus,
\begin{align*}
\frac{\partial E_{\hat{\boldsymbol{w}},\hat{\boldsymbol{z}}}\left[\ell_{\alpha,\gamma,\tau}\left(\boldsymbol{\Theta},\boldsymbol{\psi}, \boldsymbol{w}, \boldsymbol{\Omega}, \boldsymbol{z}, \boldsymbol{Y} ~|~ \boldsymbol{A}\right)\right] + \Lambda}{\partial\Theta_{k,l,p}} &=\frac{\alpha - 1}{\boldsymbol{\Theta}_{k,l,p}}
+ \sum_{(m,n) \in \boldsymbol{A}_{\cdot,\cdot,p}} \frac{\boldsymbol{w}_{mk} \boldsymbol{z}_{nl}}{\boldsymbol{\Theta}_{k,l,\boldsymbol{Y}_{m,n}}} - \lambda_{\boldsymbol{\Theta}_{k,l}}=0\\
\alpha - 1
+ \sum_{(m,n) \in \boldsymbol{A}_{\cdot,\cdot,p}} \boldsymbol{w}_{mk} \boldsymbol{z}_{nl} - \lambda_{\boldsymbol{\Theta}_{k,l}}\boldsymbol{\Theta}_{k,l,p} &= 0\\
P\alpha - P
+  \sum_{(m,n) \in \boldsymbol{A}} \boldsymbol{w}_{mk} \boldsymbol{z}_{nl} - \lambda_{\boldsymbol{\Theta}_{k,l}} &= 0\\
\boldsymbol{\Theta}_{k,l,p} &= \frac{\alpha - 1
+ \sum_{(m,n) \in \boldsymbol{A}_{\cdot,\cdot,p}}  \boldsymbol{w}_{mk} \boldsymbol{z}_{nl}}{P\alpha - P
+ \sum_{(m,n) \in \boldsymbol{A}} \boldsymbol{w}_{mk} \boldsymbol{z}_{nl}}\\
\frac{\partial E_{\hat{\boldsymbol{w}},\hat{\boldsymbol{z}}}\left[\ell_{\alpha,\gamma,\tau}\left(\boldsymbol{\Theta},\boldsymbol{\psi}, \boldsymbol{w}, \boldsymbol{\Omega}, \boldsymbol{z}, \boldsymbol{Y} ~|~ \boldsymbol{A}\right)\right] + \Lambda}{\partial \boldsymbol{\psi}_k}
&=\frac{\gamma - 1}{\boldsymbol{\psi}_k} + \sum_{m} \frac{\boldsymbol{w}_{mk}}{\boldsymbol{\psi}_k} - \lambda_{\boldsymbol{\psi}}= 0\\
\gamma - 1 + \sum_{m} \boldsymbol{w}_{mk} - \lambda_{\boldsymbol{\psi}}\boldsymbol{\psi}_k &= 0\\
K\gamma - K + M  - \lambda_{\boldsymbol{\psi}} &= 0\\
\boldsymbol{\psi}_k &= \frac{\gamma - 1 + \sum_{m} \boldsymbol{w}_{mk}}{K\gamma - K + M}\\
\end{align*}
\begin{align*}
\frac{\partial E_{\hat{\boldsymbol{w}},\hat{\boldsymbol{z}}}\left[\ell_{\alpha,\gamma,\tau}\left(\boldsymbol{\Theta},\boldsymbol{\psi}, \boldsymbol{w}, \boldsymbol{\Omega}, \boldsymbol{z}, \boldsymbol{Y} ~|~ \boldsymbol{A}\right)\right] + \Lambda}{\partial \boldsymbol{\Omega}_l}
&= \frac{\tau - 1}{\boldsymbol{\Omega}_l}  + \sum_{n} \frac{\boldsymbol{z}_{nl}}{\boldsymbol{\Omega}_l} - \lambda_{\boldsymbol{\Omega}} = 0\\
\tau - 1 + \sum_{n} \boldsymbol{z}_{nl} - \lambda_{\boldsymbol{\Omega}}\boldsymbol{\Omega}_l  &= 0\\
L\tau - L + N - \lambda_{\boldsymbol{\Omega}}  &= 0\\
\boldsymbol{\Omega}_l &= \frac{\tau - 1 + \sum_{n} \boldsymbol{z}_{nl}}{L\tau - L + N}\\
\sum_p \boldsymbol{\Theta}_{k,l,p} &= 1\\
\sum_{k=1}^{K} \boldsymbol{\psi}_{k} &= 1\\  \sum_{l=1}^{L} \boldsymbol{\Omega}_{l} &= 1
\end{align*}

\subsubsection{Belief Propagation Equations for EM}
\label{sec:belief}
Note that the hidden variables $\boldsymbol{w}$ and $\boldsymbol{z}$ have a bipartite dependency structure, where (conditioned on all of the model parameters and observed variables) the class of any hidden variable is independent of all other hidden variables, given only the hidden variables it is assigned to:
\begin{align*}
\hat{p}_{\alpha,\gamma,\tau}\left(\boldsymbol{w}_{m'}~|~\boldsymbol{\Theta},\boldsymbol{\psi},  \boldsymbol{\Omega}, (\boldsymbol{z}_n : (n,m) \in \boldsymbol{A}), \boldsymbol{Y}, \boldsymbol{A}\right)
&=p_{\alpha,\gamma,\tau}\left(\boldsymbol{w}_{m'} ~|~ \boldsymbol{\Theta},\boldsymbol{\psi}, \boldsymbol{w}_{-m'}, \boldsymbol{\Omega},  \boldsymbol{z}, \boldsymbol{Y}, \boldsymbol{A}\right)
\end{align*}
(likewise for $\boldsymbol{z}_{n'}$). According to this observation,
\begin{align*}
\MoveEqLeft
p_{\alpha,\gamma,\tau}\left(\boldsymbol{w}_{m'}= k'~|~ \boldsymbol{\Theta},\boldsymbol{\psi},  \boldsymbol{\Omega},  \boldsymbol{Y}, \boldsymbol{A}\right)\\
&=\sum_{(\boldsymbol{z}_n : n \in \boldsymbol{A}_{m',\cdot}) \in \mathbb{Z}_L^{\|\boldsymbol{A}{m',\cdot}\|}} \frac{p_{\alpha,\gamma,\tau}\left(\boldsymbol{w}_{m'}, (\boldsymbol{z}_n : n \in \boldsymbol{A}_{m',\cdot}), \boldsymbol{Y}_{m',\cdot}~|~\boldsymbol{\Theta},\boldsymbol{\psi},  \boldsymbol{\Omega},\boldsymbol{Y}_{-m',\cdot}, \boldsymbol{A}\right)}
{p_{\alpha,\gamma,\tau}\left(\boldsymbol{Y}_{m',\cdot}~|~\boldsymbol{\Theta},\boldsymbol{\psi},  \boldsymbol{\Omega},\boldsymbol{Y}_{-m',n}, \boldsymbol{A}\right)}.\\
&=\sum_{(\boldsymbol{z}_n : n \in \boldsymbol{A}_{m',\cdot}) \in \mathbb{Z}_L^{\|\boldsymbol{A}_{m',\cdot}\|}}  \frac{ p(\boldsymbol{w}_{m'}|\boldsymbol{\psi})\prod_{n \in \boldsymbol{A}_{m',\cdot}}p(\boldsymbol{Y}_{m',n}|{\boldsymbol{\Theta},\boldsymbol{w}_m,\boldsymbol{z}_n})\hat{p}_{\alpha,\gamma,\tau}\left(\boldsymbol{z}_n~|~ \boldsymbol{\Theta},\boldsymbol{\psi},  \boldsymbol{\Omega},  \boldsymbol{Y}_{-m',\cdot}, \boldsymbol{A}\right)}
{p_{\alpha,\gamma,\tau}\left(\boldsymbol{Y}_{m',\cdot}~|~\boldsymbol{\Theta},\boldsymbol{\psi},  \boldsymbol{\Omega},\boldsymbol{Y}_{m',\cdot}, \boldsymbol{A}\right)}\\
&= \frac{p(\boldsymbol{w}_{m'}|\boldsymbol{\psi})\prod_{n \in \boldsymbol{A}_{m',\cdot}}\sum_{l \in \mathbb{Z}_L} p(\boldsymbol{Y}_{m',n}|{\boldsymbol{\Theta},\boldsymbol{w}_m,\boldsymbol{z}_n})\hat{p}_{\alpha,\gamma,\tau}\left(\boldsymbol{z}_n = l ~|~ \boldsymbol{\Theta},\boldsymbol{\psi},  \boldsymbol{\Omega},  \boldsymbol{Y}_{-m',n}, \boldsymbol{A}\right)}
{p_{\alpha,\gamma,\tau}\left(\boldsymbol{Y}_{m',\cdot}~|~\boldsymbol{\Theta},\boldsymbol{\psi},  \boldsymbol{\Omega},\boldsymbol{Y}_{-m,n}, \boldsymbol{A}\right)}\\
&= p(\boldsymbol{w}_{m'}|\boldsymbol{\psi})\prod_{n \in \boldsymbol{A}_{m',\cdot}}\sum_{l \in \mathbb{Z}_L} p(\boldsymbol{Y}_{m',n}|{\boldsymbol{\Theta},\boldsymbol{w}_m,\boldsymbol{z}_n})\hat{p}_{\alpha,\gamma,\tau}\left(\boldsymbol{z}_n = l ~|~ \boldsymbol{\Theta},\boldsymbol{\psi},  \boldsymbol{\Omega},  \boldsymbol{Y}, \boldsymbol{A}\right)\\
&=  \boldsymbol{\psi}_{\boldsymbol{w}_{m'}} \prod_{n \in \boldsymbol{A}_{m',\cdot}}\sum_{l \in \mathbb{Z}_L} \boldsymbol{\Theta}_{\boldsymbol{w}_{m'},\boldsymbol{z}_n,\boldsymbol{Y}_{m',n}}\hat{p}_{\alpha,\gamma,\tau}\left(\boldsymbol{z}_n = l ~|~ \boldsymbol{\Theta},\boldsymbol{\psi},  \boldsymbol{\Omega},  \boldsymbol{Y}_{-(m',n)}, \boldsymbol{A}\right)\\
&=  \hat{p}_{\alpha,\gamma,\tau}\left(\boldsymbol{w}_{m'},  (\boldsymbol{Y}_{m',n}:n \in \boldsymbol{A}_{m',\cdot})~|~\boldsymbol{\Theta},\boldsymbol{\psi},  \boldsymbol{\Omega},(\boldsymbol{Y}_{m,n}:n \not\in \boldsymbol{A}_{m',\cdot}), \boldsymbol{A}\right)\\
\end{align*}

And so:
\begin{align*}
\MoveEqLeft
p_{\alpha,\gamma,\tau}\left(\boldsymbol{w}_{m'}= k'~|~ \boldsymbol{\Theta},\boldsymbol{\psi},  \boldsymbol{\Omega},  \boldsymbol{Y}, \boldsymbol{A}\right)\\
&= \frac{p_{\alpha,\gamma,\tau}\left(\boldsymbol{w}_{m'} = k',  (\boldsymbol{Y}_{m',n}:n \in \boldsymbol{A}_{m',\cdot})~|~\boldsymbol{\Theta},\boldsymbol{\psi},  \boldsymbol{\Omega}, \boldsymbol{A},(\boldsymbol{Y}_{m,n}:n \not\in \boldsymbol{A}_{m',\cdot})\right)}{\sum_{k \in \mathbb{Z}_K} p_{\alpha,\gamma,\tau}\left(\boldsymbol{w}_{m'} = k,  (\boldsymbol{Y}_{m',n}:n \in \boldsymbol{A}_{m',\cdot})~|~\boldsymbol{\Theta},\boldsymbol{\psi},  \boldsymbol{\Omega}, \boldsymbol{A},(\boldsymbol{Y}_{m,n}:n \not\in \boldsymbol{A}_{m',\cdot})\right)}\\
&= \frac{\boldsymbol{\psi}_{\boldsymbol{w}_{m'}} \prod_{n \in \boldsymbol{A}_{m', \cdot}}\sum_{l \in \mathbb{Z}_L} \boldsymbol{\Theta}_{\boldsymbol{w}_m,\boldsymbol{z}_n,\boldsymbol{Y}_{m',n}}\hat{p}_{\alpha,\gamma,\tau}\left(\boldsymbol{z}_n = l ~|~ \boldsymbol{\Theta},\boldsymbol{\psi},  \boldsymbol{\Omega},  \boldsymbol{Y}_{-(m',n)}, \boldsymbol{A}\right)}
{\sum_{k \in \mathbb{Z}_K}\boldsymbol{\psi}_{k} \prod_{n \in \boldsymbol{A}_{m', \cdot}}\sum_{l \in \mathbb{Z}_L} \boldsymbol{\Theta}_{k,\boldsymbol{z}_n,\boldsymbol{Y}_{m',n}}\hat{p}_{\alpha,\gamma,\tau}\left(\boldsymbol{z}_n = l ~|~ \boldsymbol{\Theta},\boldsymbol{\psi},  \boldsymbol{\Omega},  \boldsymbol{Y}_{-(m',n)}, \boldsymbol{A}\right)}\\
\end{align*}

Note that this requires us to use $\hat{p}_{\alpha,\gamma,\tau}\left(\boldsymbol{z}_n = l ~|~ \boldsymbol{\Theta},\boldsymbol{\psi},  \boldsymbol{\Omega},  \boldsymbol{Y}_{-(m',n)},\boldsymbol{A}\right)$ for each $n \in \boldsymbol{A}_{m'}$. Anticipating its use the analogous message passing formula for $\boldsymbol{z}_{n'}$, we can calculate $\hat{p}_{\alpha,\gamma,\tau}\left(\boldsymbol{w}_{m'} = k ~|~ \boldsymbol{\Theta},\boldsymbol{\psi},  \boldsymbol{\Omega},  \boldsymbol{Y}_{-(m',n')},\boldsymbol{A}\right)$ for each $n' \in \boldsymbol{A}_{m'}$ by removing the $n'$th term in each of the products above.

\section{A Neural Probabilistic Model for Estimating Label Distributions}
\label{sec:neural_model}
\begin{wrapfigure}{r}{.5\textwidth}
    \includegraphics[width=.445\textwidth]{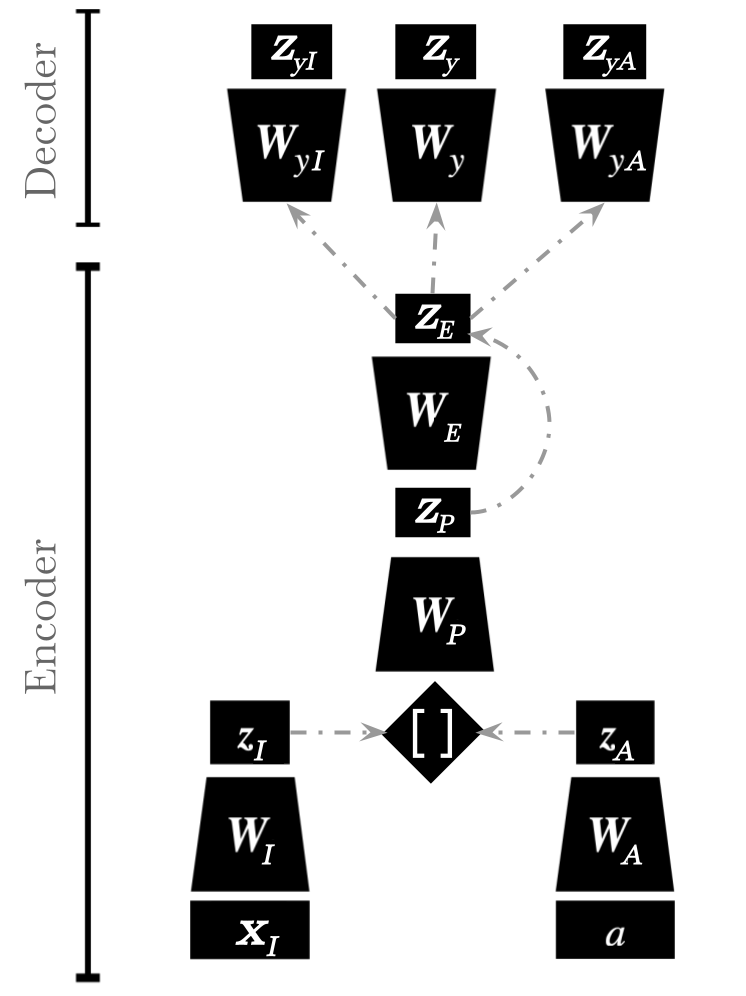}
    \caption{Block diagram showing the main components/parameters of the proposed LDL-NM.}
    \label{fig:my_label}
\end{wrapfigure}
The design of our proposed neural model was inspired/motivated by the structure of the graphical models presented in the last section.
Specifically, we envisioned the model as forming a type of generative process that could generate item label, annotator label, and full ground-truth distributions over classes. We furthermore opted to craft an encoder-decoder design for the underlying artificial neural network (ANN) in order to facilitate tractable inference and parameter learning.

Our probabilistic neural model (LDL-NM) is defined by a set of synaptic weight parameter matrices housed in two constructs $\Theta_e= \{\mathbf{W}_I, \mathbf{W}_A,  \mathbf{W}_P, \mathbf{W}_E\}$ and $\Theta_d = \{\mathbf{W}_{yI}, \mathbf{W}_{yA}, \mathbf{W}_y \}$ (bias vectors omitted for clarity), where $\Theta_e$ contains the encoder parameters and $\Theta_d$ contains the decoder parameters. 
The model is designed, for each data item feature vector $\mathbf{x}_i$ and annotator (index) $a$ pair, to estimate the values of a set of target label distributions, i.e., the item label distribution $\mathbf{y}_I$ the annotator label distribution $\mathbf{y}_A$, and the (ground-truth) label distribution $\mathbf{y}$. 

The output of the encoder is the latent representation of data items and annotators -- note that the data item is projected to the space $\mathbf{z}_I$ while the annotator identification integer is embedded into the space $\mathbf{z}_A$.
As a result, the encoder, which takes in as input the data item feature vector $\mathbf{x}_I \in \mathcal{R}^{J \times 1}$ ($J$ is the dimensionality of the feature space) and the annotator identifier $a \in \mathbb{N}_0$ (the set of natural numbers including zero), computes the following output:
\begin{align}
    \mathbf{z}_I &= \mathbf{W}_I \cdot \mathbf{x}_I, \; \mbox{and}, \; \mathbf{z}_A = \mathbf{W}_A \cdot \text{enc}(a) \label{eq:encode_lat}\\
    \mathbf{z}_P &=  \phi(\mathbf{W}_P \cdot \phi([\mathbf{z}_I, \mathbf{z}_A])), \; \mbox{and}, \; \mathbf{z}_E =  \phi\big((\mathbf{W}_E \cdot \mathbf{z}_P) + \mathbf{z}_P\big) \label{eq:encode_out}
\end{align}
where $\cdot$ denotes matrix multiplication, $[\mathbf{a},\mathbf{b}]$ represents vector combination operation applied to input vectors $\mathbf{a}$ and $\mathbf{b}$ (such as concatenation or element-wise summation), and $\text{enc}(a)$ is a function to convert integer $a$ to a corresponding one-hot encoding binary vector. 
$\phi(\mathbf{v}) = 1 / |1 + \mathbf{v}|$ is the softsign elementwise activation function. $\mathbf{z}_I \in \mathcal{R}^{J_I \times 1}$ and $\mathbf{z}_A \in \mathcal{R}^{J_A \times 1}$ where $J_I$ and $J_A$ are their respective embedding dimensionalities.
An additional linear projection is applied to the combined item and annotator embeddings via matrix $\mathbf{W}_P$ to reduce the dimesionality further to $\mathbf{z}_E \in \mathcal{R}^{J_P \times 1}$ before running the representation through one more non-linear transform (to obtain encoder output $\mathbf{z}_E$). 

The decoder, which takes in as input the latent code produced by the encoder $\mathbf{z}_E$, computes its outputs (three different label distribution estimates) as follows:
\begin{align}
    \mathbf{z}_{y} &= \sigma(\mathbf{W}_y \cdot \mathbf{z}_E), \;
    \mathbf{z}_{yI} = \sigma(\mathbf{W}_{yI} \cdot \mathbf{z}_E), \; 
    \mathbf{z}_{yA} = \sigma(\mathbf{W}_{yA} \cdot \mathbf{z}_E) \label{eq:decode_out}
\end{align}
where $\sigma(\mathbf{v}) = \exp(\mathbf{v})/\sum_j \exp(\mathbf{v})[j]$ is the softmax function ($\mathbf{v}[j]$ retrieves the $j$the value/element of the vector $\mathbf{v}$). Notice that a residual connection has been introduced in Equation \ref{eq:encode_out} to improve gradient flow during model training. 
Note that, under our proposed LDL-NM, $\mathbf{z}_y$ is interpreted as $p(\mathbf{y}|\mathbf{x}_i, a)$, $\mathbf{z}_{yI}$ is interpreted as $p(\mathbf{y}_I|\mathbf{x}_i, a)$, and $\mathbf{z}_{yA}$ is interpreted as $p(\mathbf{y}_A|\mathbf{x}_i, a)$.


\subsection{Parameter Optimization}
\label{sec:optimization}

To train model parameters $\Theta = \{ \Theta_e, \Theta_d \}$, we propose the following multi-objective function:
\begin{align}
    \mathcal{L}(\Theta) = -\sum_j \big( \mathbf{y} \otimes \log( \mathbf{z}_{y}) \big)[j] + \text{KL}(\mathbf{y}_I||\mathbf{z}_{yI}) + \text{KL}(\mathbf{y}_A||\mathbf{z}_{yA}) \label{eq:ldlnn_cost}
\end{align}
where the first term is the negative Categorical log likelihood of the target one-hot encoded label $\mathbf{y}$ and the second and third terms measure the Kullback-Leibler (KL) divergence of between the decoder's estimate of the item label distribution $\mathbf{y}_i$ and annotator label distribution $\mathbf{y}_a$, respectively. Specifically, the form of the KL divergence we use compares two multinomial/multinoulli distributions\footnote{The forms of these KL divergences were derived by exploiting Stirling's approximation \cite{jaynes2003probability} to deal with the factorial that appears in the definition of the multinomial likelihood. Full derivation included in Appendix\ref{sec:kl_deriv_ldlnm}. }:
\begin{align}
    -\text{KL}(\mathbf{y}_I||\mathbf{z}_{yi};\Theta) &= -\sum_j \big( \mathbf{y}_I \otimes \log \mathbf{y}_I \big)[j] + \sum_j \big( \mathbf{y}_I \otimes \log \mathbf{z}_{yI} \big)[j] \label{eq:kl_item} \\
    -\text{KL}(\mathbf{y}_A||\mathbf{z}_{yA};\Theta) &= -\sum_j \big( \mathbf{y}_A \otimes \log \mathbf{y}_A \big)[j] + \sum_j \big( \mathbf{y}_A \otimes \log \mathbf{z}_{yA} \big)[j]
\end{align}
where $\otimes$ denotes the Hadamard product and unit of measurement for each KL divergence is the nat. The LDL-NM parameters are adjusted to minimize the function defined in Equation \ref{eq:ldlnn_cost} by calculating the gradients with respect to both the encoder and decoder weights, i.e., $\frac{\partial \mathcal{L}(\Theta_e,\Theta_d)}{\partial \Theta_e}$ and $\frac{\partial \mathcal{L}(\Theta_e,\Theta_d)}{\partial \Theta_d}$. The resultant partial derivatives are then used to change the current values in $\Theta_e$ and $\Theta_d$ via stochastic gradient descent or with a more advanced adaptive learning rate rule such as Adam \cite{kingma2014adam}.

\subsection{Test-Time Usage}
\label{sec:test_time_inf}

While the LDL-NM is a rather general-purpose neural model and can be used in a variety of interesting ways when it comes to modeling items and annotators, in this study, there are two particular use-cases of the LDL-NM we are interested in.

\begin{algorithm}[!t]
\caption{The iterative inference process for the LDL-NM.}
\label{algo:inf_process}
\begin{algorithmic}
   \State {\bfseries Input:} Sample $(\mathbf{x}_I,\mathbf{y}_I)$, $\beta$, $\gamma$, $K$, $\sigma_a$, and $\Theta = \{\Theta_e,\Theta_d\}$
   \Function{InferA}{$\mathbf{x}_I, \mathbf{y}_I, \Theta$} \Comment Run $K$-step inference to find $\mathbf{z}_A$
        \State $\mathbf{z}_A \sim \mathcal{N}(\mu=0,\sigma_a)$, \; or, 
        \; $\mathbf{z}_A = \mathbf{0}$\Comment{Initialize annotator embedding(s)}
        \For{$k = 0$ to $K$}
            \State Run Equations \ref{eq:encode_lat}-\ref{eq:encode_out} \; and then \; $\mathbf{z}_{yI} = \sigma(W_{yI} \cdot \mathbf{z}_E)$
            \State Calculate $\text{KL}(\mathbf{y}_I||\mathbf{z}_{yI};\Theta)$
            \If{Termination criterion not met}
                \State $\mathbf{z}_A \gets \mathbf{z}_A - \beta \frac{\partial \text{KL}(\mathbf{y}_I||\mathbf{z}_{yI};\Theta)}{\partial \mathbf{z}_A} - \gamma \mathbf{z}_A$
            \EndIf
        \EndFor
        \State \textbf{Return} $\mathbf{z}_A$ and  $\text{KL}(\mathbf{y}_I||\mathbf{z}_{yI};\Theta)$
    \EndFunction
\end{algorithmic}
\end{algorithm}

The first involves using the model to make a prediction of the label for item feature vector $\mathbf{x}_i$ given only its particular item label distribution  $\mathbf{y}_i$. In this case, the LDL-NM makes multiple predictions of the possible label, one per annotator embedding stored in its internal memory.
This inference is done by first modifying Equation \ref{eq:encode_out} as follows:
\begin{align}
    \mathbf{z} = \phi(\mathbf{W}_A + \mathbf{z}_I), \; \mbox{and}, \; \mathbf{z}_e =  \phi\big((\mathbf{W}_E \cdot \mathbf{z}) + \mathbf{z}\big) \label{eq:ensemble_decode}
\end{align}
where $\mathbf{z}_I$ is added to each column of $\mathbf{W}_A$. When using Equation \ref{eq:ensemble_decode}, the resulting outputs $\mathbf{z}_y$, $\mathbf{z}_{yi}$, and $\mathbf{z}_{ya}$ will each contain $A$ columns (one per annotator) so in order to compute the final prediction for each distribution, one could compute the expectation across columns or compute the $\arg\max$ of each column and compute the mode across the full set of guessed class integers.

The second case involves conducting iterative inference to find an annotator embedding(s) when no annotator identifier is provided, i.e., the model is given only the item feature vector and item label distribution. Specifically, the iterative inference process proceeds as depicted in Algorithm \ref{algo:inf_process}. Note that $\gamma$ is a decay factor to smooth and regularize the values of the  resulting embeddings.

\section{Experiments}
\label{sec:experiments}
\subsection{Data}
\label{sec:data}

\begin{table}
\centering
    \begin{tabular}{ccccc}
    \toprule
         &\# annotators & \# label  & mean  &\# of 
         \\
        dataset & per item & classes & entropy &annotators
        \\\hline
         JQ1 \cite{P16-1099}   & 10 & 5 & $0.746$ & 1185 \\
         JQ2 \cite{P16-1099}  & 10 & 5 & $0.586$ & 1185\\ 
         JQ3 \cite{P16-1099}  & 10 & 12 & $0.993$ & 1185\\
         \bottomrule
    \end{tabular}
    \caption{Summary of datasets on which we conduct our experiments. Each of these contain 2,000 items.}\vspace{-.5cm}
    \label{tab:dataset}
\end{table}

We conducted our experiments on publicly available human annotated datasets. 
Each dataset consists of 2,000 social media posts and employs a 50/25/25 percent for train/dev/test split.

Liu et al \cite{P16-1099} asked five annotators each from MTurk and FigureEight to label work related tweets according to three questions and associated multiple choice responses: point of view of the tweet (\textbf{JQ1}: \emph{1st person}, \emph{2nd person}, \emph{3rd person}, \emph{unclear}, or \emph{not job related}), subject's employment status (\textbf{JQ2}: \emph{employed}, \emph{not in labor force}, \emph{not employed}, \emph{unclear}, and \emph{not job-related}), and employment transition event (\textbf{JQ3}: \emph{getting hired/job seeking}, \emph{getting fired}, \emph{quitting a job}, \emph{losing job some other way}, \emph{getting promoted/raised}, \emph{getting cut in hours}, \emph{ complaining about work}, \emph{offering support}, \emph{going to work}, \emph{coming home from work}, \emph{none of the above but job related}, and \emph{not job-related}).

We train and test on the following models. For natural language data, in order to generate item feature vectors ($\mathbf{x}_i$) for LDL-NM, we used sentence embeddings from \cite{reimers-2019-sentence-bert} using the pretrained \texttt{paraphrase-MiniLM-L6-v2} model. 

\begin{description}[leftmargin=0cm]
\item[PD] is a baseline supervised model with no unsupervised graphical model. It is The supervised model (CNN) is a 1D convolutional neural network \cite{kim2014convolutional}, with three convolution/max pool layers followed by a dropout and softmax layer, implemented in TensorFlow \cite{tensorflow2015-whitepaper}. It uses KL divergence for the cost function. The model has a 20,000-word lexicon at the input layer, transforms each word into a 100-dimension vector via the Glove 2B-tweet corpus \cite{pennington2014glove}, and admits text sequences of up to 1,000 tokens per data item.
 
\item[FMM] is the baseline model, with the best-performing graph-based model from Weerasooriya et al. \cite{Weerasooriya2020} used as a guiding model, in a manner analogous to how the graph model introduced in this paper is used. The main difference between their model and ours is that it only performs item label distribution clustering; there are no annotator clusters.
\item[PGM] is the proposed baseline supervised Bayesian model, with the graph model introduced here for guidance. We set all of the Dirichlet parameters $\alpha$, $\gamma$, and $\tau$ to $2$. We consider two different learning algorithms: simulated annealing (with temperature schedule $T(t) = 1/(t+1)$) and expectation maximization (EM) with belief propagation.
\item[LDL-NM] is the neural model described earlier. The item ($\mathbf{z}_I$) and annotator ($\mathbf{z}_A$) embeddings were combined by setting the $[\mathbf{a},\mathbf{b}]$ to be vector concatenation. We furthermore regularized model parameters during training by applying drop-out, with a drop probability of $p = 0.5$, to the output activity of $\mathbf{z}_P$ and $\mathbf{z}_E$. The Adam adaptive learning rate \cite{kingma2014adam} was used to optimize parameters by using gradients calculated over mini-batches of $256$ samples for $200$ epochs. Model parameters were initialized from random orthogonal matrices \cite{saxe2013exact}.
\end{description}

For each of the the graphical models and LDL-NM we performed parameter search on the number of item and annotator clusters (or, analogously, the hidden layer sizes of the item and annotator encoders) $K, L \in \{3, \ldots, 20\}$ and report the results on the best performing model on development data.
We evaluate these models using two different metrics. To evaluate label distribution prediction we report over the test set the mean KL divergence between each gold standard label distribution and the predicted label distribution $\mathrm{KL}(h_{\rm dist}(x) \| y)$. To evaluate single label prediction, we report the accuracy measured over the test set. 

Our experiments were conducted on; \#1 - Local machine with Intel i6-7600k (4 cores) at 4.20GHz, 32GB RAM, and nVidia GeForce RTX 2070 Super 8GB VRAM and \#2 Internal cluster with Intel(R) Xeon E7 v4, 264GB RAM, and GPU Tesla P4 8GB. Our worst case scenario was machine \#1 in our setup and dataset JQ3. The runtime for a single pass of experiments on a single dataset for PD took 2 minutes, FMM took 2 hours, PGM (annealing) took 9 hours, PGM (BP) took 7 hours, and the LDL-NM took 1 hour.


\subsection{Results}
\label{sec:results}
Table \ref{tab:main_results} shows the main results. We first note that all models outperform PD, i.e., supervised learning with no clustering. For RQ1, both PGM models slightly underperform FMM on both tasks and on all datasets. Since the PGM can be seen as a finer-grained version of FMM (i.e., one that clusters in annotator space in addition to item space), we suspect this is due to overfitting that we have, thus far been, unable to control. 
For RQs 2 and 3, LDL-NM outperforms all other models, including PD and all models with graphical clustering. Additionally, the neural model has the advantage of running as a single stage process, without the unsupervised clustering stage that the graphical models require (and which is often much slower than backpropagation). 

We repeated experiments for LDL-NM to include error bars for KL-divergence and accuracy, included in Table~\ref{tab:main_results}. In case of PD, FMM, PGM (annealing), and PGM (BP), final supervised learning classification piece was repeated 100 times (trained and evaluated) to calculate error bars. For LDL-NM, evaluation piece was repeated 100 times by adding dropout layer for analysis.

\begin{table}
  \caption{Experimental parameters of our unsuperivsed learning models. $K$ is the number of item classes, $L$ is the number of annotator classes, and $D_{KL}$ is the KL divergence when evaluated against empirical ground truth. *FMM model is clustered only on item classes.}
  \label{tab:unsupervised_results}
  \centering
  \begin{tabular}{lcccc}
    \toprule
    \cmidrule(r){2-5}
    Dataset     &  & FMM* &  PGM (Annealing) & PGM (BP)\\
    \midrule
    JQ1 & $D_{KL}$ & 0.193 & 0.651 & 0.523 \ \\
        & $L$ & - & 4 & 12 \\
        & $K$ & 14 & 4 & 10 \\ \midrule
    JQ2 & $D_{KL}$ & 0.170 & 0.889 & 0.906 \ \\
        & $L$ & - & 5 & 12 \\
        & $K$ & 7 & 6 & 8 \\ \midrule
    JQ3 & $D_{KL}$ & 0.269 & 1.389 & 1.190 \ \\
        & $L$ & - & 9 & 11 \\
        & $K$ & 35 & 14 & 12 \\ 

    \bottomrule
  \end{tabular}
\end{table}

\begin{table}
  \caption{Experimental results for classification task. New methods (PGM, LDL-NM) using development set for each dataset. PD is a baseline where only a CNN classification is run and FMM is clustering only on the labels. The predictions are compared against the empirical ground truth.}
  \label{tab:main_results}
  \centering
  \begin{tabular}{lccccc}
    \toprule
    & \multicolumn{5}{c}{KL-Divergence}                   \\
    \cmidrule(r){2-6}
    Dataset     & PD    & FMM &  PGM (Annealing) & PGM (BP) &LDL-NM \\
    \midrule
    JQ1 & 1.092$\pm$0.004  & 0.460$\pm$0.001 & 0.652$\pm$0.005  & 0.538$\pm$0.010  & 0.408$\pm$0.004 \\
    JQ2 & 1.088$\pm$0.003 & 0.514$\pm$0.002 & 0.884$\pm$0.004 & 0.624$\pm$0.017& 0.501$\pm$0.024 \\
    JQ3 & 1.462$\pm$0.004 & 0.888$\pm$0.001 & 1.201$\pm$0.005 &  0.951$\pm$0.016 &0.864$\pm$0.001 \\

    \bottomrule
    \bottomrule
     & \multicolumn{5}{c}{Accuracy}   \\
     \midrule
     JQ1 & 0.494$\pm$0.001 & 0.892$\pm$0.000& 0.730$\pm$0.000 &  0.727$\pm$0.007& 0.915$\pm$0.001 \\
     JQ2 & 0.475$\pm$0.001 & 0.873$\pm$0.001 & 0.579$\pm$0.041 &  0.663$\pm$0.013& 0.910$\pm$0.001 \\
     JQ3 & 0.284$\pm$0.020 & 0.880$\pm$0.001 & 0.290$\pm$0.002 &  0.250$\pm$0.007 & 0.923$\pm$0.0015 \\
     \bottomrule
  \end{tabular}
  \vspace{-.5cm}
\end{table}

\section{Qualitative Results for LDL-NM: t-SNE Visualization}
\label{sec:tsne_ldlnm}
\begin{figure}[!ht]
    \centering
    \includegraphics[width=0.45\textwidth]{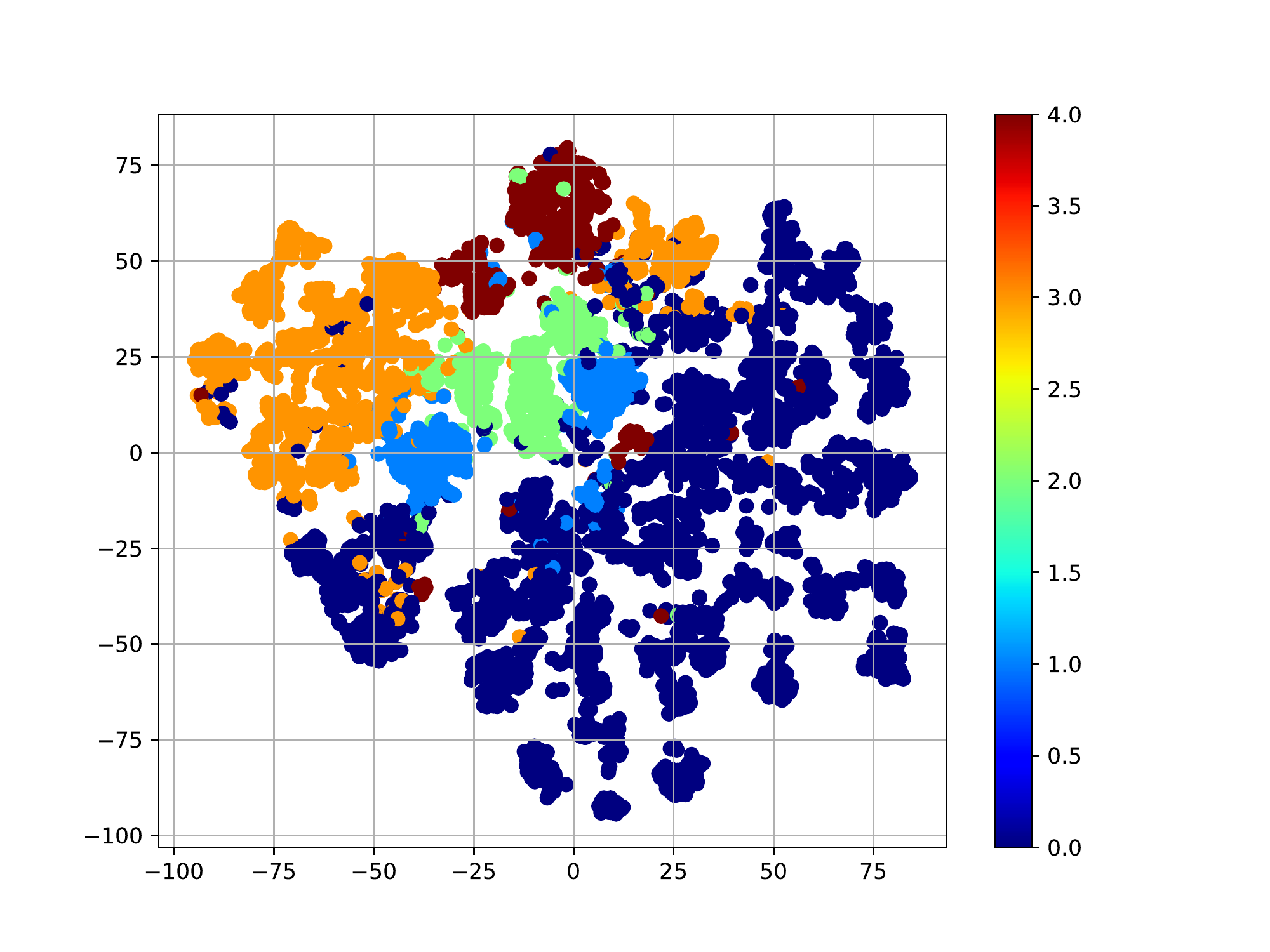}
    \includegraphics[width=0.45\textwidth]{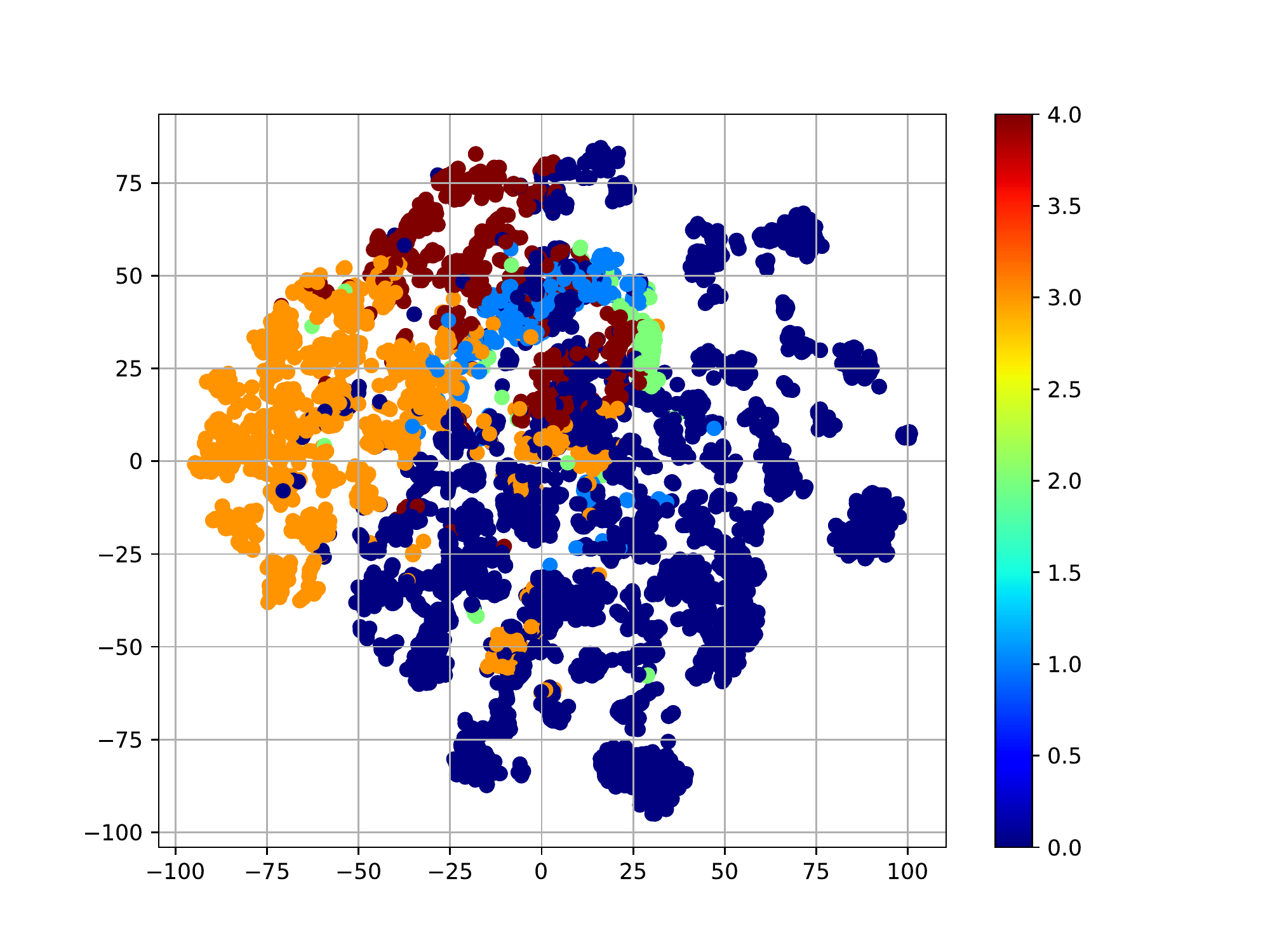}
    \includegraphics[width=0.45\textwidth]{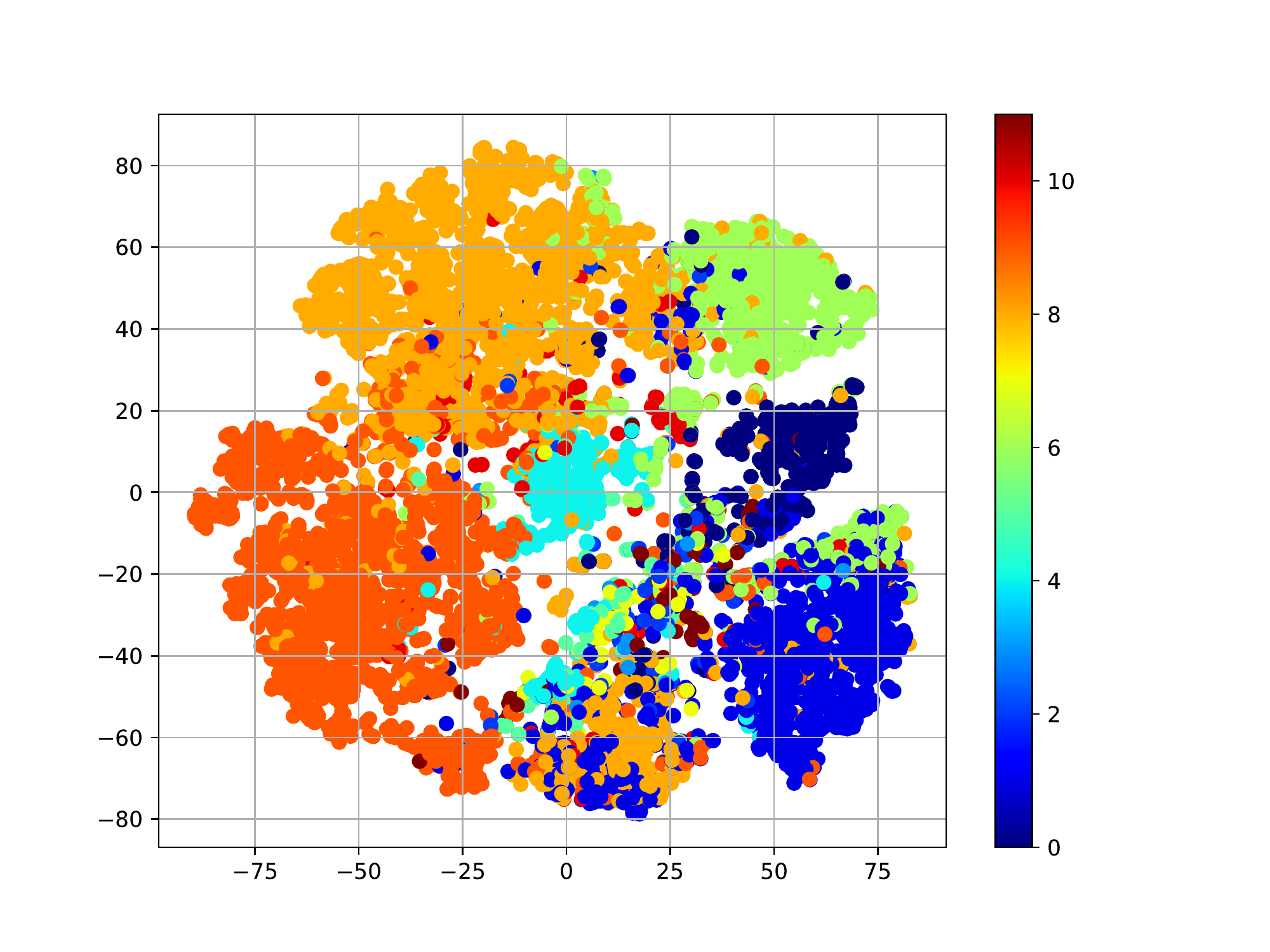}
    \caption{t-SNE plot for training set from LDL-NM models. Each color represents a label class. (Top Left) Plot for JQ1 dataset, which has five label classes ($C$). (Top Right) Plot for JQ2 dataset, which also has five label classes ($C$). (Bottom) Plot for JQ3 dataset, which has 12 label classes ($C$). }
    \label{fig:ts}
\end{figure}

We generated t-SNE visualizations for our experiments using the \textbf{LDL-NM} models, plots show label class separation. Label classes for Jobs dataset by \cite{P16-1099}, point of view of the tweet (\textbf{JQ1}: \emph{1st person}, \emph{2nd person}, \emph{3rd person}, \emph{unclear}, or \emph{not job related}), subject's employment status (\textbf{JQ2}: \emph{employed}, \emph{not in labor force}, \emph{not employed}, \emph{unclear}, and \emph{not job-related}), and employment transition event (\textbf{JQ3}: \emph{getting hired/job seeking}, \emph{getting fired}, \emph{quitting a job}, \emph{losing job some other way}, \emph{getting promoted/raised}, \emph{getting cut in hours}, \emph{ complaining about work}, \emph{offering support}, \emph{going to work}, \emph{coming home from work}, \emph{none of the above but job related}, and \emph{not job-related}). Figures are included in Figure~\ref{fig:ts}.

\section{Discussion}
\label{sec:discussion}
\paragraph{Limitations.} We would have liked to test our model on more datasets, however, there are very few multiclass, publicly available datasets that include information about annotator assignments. This information is either discarded or withheld. Without annotator assignments, it is impossible to perform clustering. We hope that these results encourage more researchers to share this information (though see caveats below).

In our graphical models we could have performed more extensive parameter search on the Dirichlet hyperparameters. Although early in our research we explored using the graphical model to predict soft clusters, we found that we obtain better results with hard clusters. Nonetheless, a soft inference model is probably a better for the purpose of comparing performance to the neural model, because the neural model is effectively performing soft prediction in it internal layers.  

Although we directly compared our model's performance to those of Weearsooriya et al. \cite{Weerasooriya2020}, who represent clustering in item label space only, we did not perform head-to-head comparisons to Venanzi et al's \cite{venanzi2014community} model, which represents clustering in annotator label space only. This is due in part to do the data from their studies no longer being available. We nonetheless intend to run their models our the data we do have in our next study.

\paragraph{Ethics review.} Before conducting the research described here, we consulted with our institutional review board(s). They determined it did not constitute human subjects research, primarily because the data was publicly available and secondary. Beyond that, all authors have basic training on conducting human subjects research from CITI\footnote{\url{https://about.citiprogram.org/en/series/human-subjects-research-hsr/}}. Moreover, we do not report on any of the details of the data we used.

\paragraph{Potential societal impact.} Our involvement in label distribution learning came from a community based participatory research group we belonged to on the use of AI technology 
in vulnerable communities as a means of preserving in AI pipelines minority perspectives that would otherwise be erased when annotator disagreement is resolved, usually in favor of the plurality label, as is common practice today. We believe these methods, coupled with demographic information on annotators and reliable confidence estimates, can lead to annotated data that is more representative of the true values within a society.

However, all statistical methods are double-edged swords. Used dishonestly, these methods could be used to misrepresent social values and opinions. Moreover, while these methods would be more informative with demographic information on the annotators, that incurs on the privacy of annotators, a group of workers who are often treated unfairly. 

\section{Conclusion}
\label{sec:conclusion}
We introduced to two new models for improving the quality of annotator labels, both from the perspective of the conventional problem of predicting the most common label and the emerging problem of predicting the distribution of labels provided. Our methods combine label distribution learning with clustering jointly in the item and annotator label distribution spaces. Our first model was an unsupervised-based graphical model that can be used as a black-box guiding model for supervised learning by replacing the given small-sample labels with ones designed to better capture the ground truth label distributions, which are unobserved. 
The second model was a neural model designed to capture internally the main features of the graphical model and provide an all-in-one supervised model that can be reconfigured to infer various aspects of the problem domain.

\bibliographystyle{alpha}
\bibliography{biblio}
\appendix
\section{Deriving the Kullback-Leibler Divergence Terms for the LDL-NM}
\label{sec:kl_deriv_ldlnm}

If we were to perform an experiment, measuring a discrete variable, and take $N$ measurements, we would obtain a histogram $c = \{c_i\}$ where $N = \sum_i c_i$ (we measures the number of times that we see each possible value that the discrete variable could take on). If $N \rightarrow \infty$ (the experiment is infinitely long), then the normalized histogram counts $c_i/N$ approach $p_i = \frac{c_i}{N}$ (a target distribution).
If we had a distribution $q$, we seek to answer the question: What is the probability of observing the histogram counts $c$ if $q$  generated the observations?\footnote{For $C$ distinct, discrete values (or classes), $p = \mathbf{p} = \{p_1, p_2,...,p_C\}$ and $q = \mathbf{q} = \{q_1, q_2,...,q_C\}$.}

We start from the defnition of the multinomial likelihood \cite{cover1999elements,shlens2006structure}, which expresses the probability of observing a histogram $c = \{c_i\}$ given that model $q = \{q_i\}$ is true:
\begin{align}
    \mathcal{F}(c|q) = \frac{N!}{\Pi_i c_i!}\Pi_i (q_i)^{c_i} \label{eq:multi_likeli}
\end{align}
where $\frac{N!}{\Pi_i c_i!}$ is the normalization constant that counts the number of combinations that could give rise to histogram $c$ ($N = \sum_i c_i$ is the total number of measurements).
To compute the joint probability of all measurements under the multinomial likelihood, we ultimately multiply the independent observations (of the histogram) together thus resulting in the geometric mean of the multinomial likelihood $\mathcal{F}(c|q)^{\frac{1}{N}}$ (which is an invariant likelihood across histogram counts). We then take the natural logarithm of the average likelihood and plug in Equation \ref{eq:multi_likeli} to obtain:
\begin{align}
    \log \mathcal{F}(c|q)^{\frac{1}{N}} &= \frac{1}{N} \log \bigg(  \frac{N!}{\Pi_i c_i!}\Pi_i (q_i)^{c_i} \bigg) \\
    &= \frac{1}{N} \log N! - \frac{1}{N} \sum_i \log c_i! + \sum_i \frac{c_i}{N} \log q_i \mbox{.} \label{eq:avg_log_like}
\end{align}
To remove the factorial in the above expression, we utilize Stirling's numerical approximation for factorials, i.e., $\log N! \simeq N \log N - N $, an approximation that becomes quite good for reasonably large $N$ \cite{dutka1991early}. Replacing all of the $\log N!$ in Equation \ref{eq:avg_log_like} with this approximation results in:
\begin{align}
    \log \mathcal{F}(c|q)^{\frac{1}{N}} &= \frac{1}{N} \big(N \log N - N) - \frac{1}{N} \sum_i (c_i \log c_i - c_i) + \sum_i \frac{c_i}{N} \log q_i \label{eq:kl_step1}\\
    &= \log N - \sum_i \frac{c_i}{N} \log c_i + \sum_i \frac{c_i}{N} \log q_i \label{eq:kl_step2}\\ 
    &= \sum_i \frac{c_i}{N} \log N - \sum_i \frac{c_i}{N} \log c_i + \sum_i \frac{c_i}{N} \log q_i \label{eq:kl_step3}
\end{align}
where the bottom expression is obtained by noting that the coefficient of $1$ that multiplies the first term $\log N$ (in Equation \ref{eq:kl_step2}) can be re-written as $\frac{N}{N} = \frac{1}{N} \sum_i c_i =  \sum_i \frac{c_i}{N}$. The derivation then continues with the applications of the rules of logarithms:
\begin{align}
    &= \sum_i \Big( \frac{c_i}{N} \log N - \frac{c_i}{N} \log c_i \Big) + \sum_i \frac{c_i}{N} \log q_i \label{eq:kl_step4} \\
    &= \sum_i \Big( -\frac{c_i}{N} \big( -\log N + \log c_i \big) \Big) + \sum_i \frac{c_i}{N} \log q_i  \label{eq:kl_step5} \\
    &= -\sum_i \Big( \frac{c_i}{N} \big( \log \frac{1}{N} + \log c_i \big) \Big) + \sum_i \frac{c_i}{N} \log q_i  \label{eq:kl_step6} \\
    &= -\sum_i \frac{c_i}{N} \log \frac{c_i}{N} + \sum_i \frac{c_i}{N} \log q_i  \label{eq:kl_step7} \mbox{.}
\end{align}
Note that, as described at the start of this section, when $N \rightarrow \infty$ then $p_i \equiv \frac{c_i}{N}$, we finally arrive at the (negative) Kullback-Leibler (KL) divergence:
\begin{align}
    &= -\sum_i p_i \log p_i + \sum_i p_i \log q_i = -\text{KL}(p||q) \mbox{.} \label{eq:kl_final}
\end{align}
This result implies that KL divergence in our case is the negative logarithm of the (geometric) mean of the multinomial likelihood (the first term of Equation \ref{eq:kl_final} is also known as the entropy of the target distribution $p$ \cite{jaynes2003probability}). Since the LDL-NM model proposed in this paper already produces the values $q_i$ for each $i$th dimension of the count histogram and we already have an estimated target $p_i$ from the annotator and/or item distributions (which are normalized to lie in the range of $[0,1]$ in the pre-processing stage, thus mimicking the required $p_i \equiv \frac{c_i}{N}$), we may directly utilize the negative of the result in Equation \ref{eq:kl_final}. Note that to recover the exact form of the KL terms presented in the main paper, one must, for $C$ classes, set either $\mathbf{p} = \mathbf{y}_{I}$ or $\mathbf{p} = \mathbf{y}_{A}$ and either $\mathbf{q} = \mathbf{z}_{yI}$ or $\mathbf{q} = \mathbf{z}_{yA}$.
\end{document}